\documentclass[journal]{IEEEtran}

\usepackage{graphicx}
\usepackage{comment}
\usepackage{amsmath,amssymb} 
\usepackage{color}

\usepackage{amsmath}
\usepackage{amssymb}

\usepackage{latexsym}
\usepackage{amsmath}
\usepackage{booktabs} 
\usepackage{subfigure}
\usepackage{multirow}

\begin{document}
%
\title{Decomposing Generation Networks with Structure Prediction for Recipe Generation}
%
%
%

\author{Hao~Wang,
        Guosheng~Lin,
        Steven~C.~H.~Hoi,~\IEEEmembership{Fellow,~IEEE}
        and~Chunyan~Miao
\thanks{Hao Wang, Guosheng Lin and Chunyan Miao are with School of Computer Science and Engineering, Nanyang Technological University; e-mail: \{hao005,gslin,ascymiao\}@ntu.edu.sg.}
\thanks{Steven C. H. Hoi is with Singapore Management University; e-mail: chhoi@smu.edu.sg.}
}

%
%

\markboth{}%
{Shell \MakeLowercase{\textit{et al.}}: Bare Demo of IEEEtran.cls for IEEE Journals}
%



\maketitle

\begin{abstract}
Recipe generation from food images and ingredients is a challenging task, which requires the interpretation of the information from another modality. Different from the image captioning task, where the captions usually have one sentence, cooking instructions contain multiple sentences and have obvious structures. To help the model capture the recipe structure and avoid missing some cooking details, we propose a novel framework: Decomposing Generation Networks (DGN) with structure prediction, to get more structured and complete recipe generation outputs. Specifically, we split each cooking instruction into several phases, and assign different sub-generators to each phase. Our approach includes two novel ideas: (i) learning the recipe structures with the global structure prediction component and (ii) producing recipe phases in the sub-generator output component based on the predicted structure. Extensive experiments on the challenging large-scale Recipe1M dataset validate the effectiveness of our proposed model, which improves the performance over the state-of-the-art results.
\end{abstract}


%
\IEEEpeerreviewmaketitle

\section{Introduction}
Due to food is very close to people's daily life, food-related research, such as food image recognition \cite{matsuda2012recognition,bossard2014food}, cross-modal food retrieval \cite{salvador2017learning,carvalho2018cross,wang2019learning} and recipe generation \cite{bosselut2017simulating,salvador2019inverse,chandu2019storyboarding}, has raised great interests recently. From a technical perspective, jointly understanding the multi-modal food data \cite{salvador2017learning} including food images and recipes remains an open research task. In this paper, we try to approach the problem of generating cooking instructions (recipes) conditioned on food images and ingredients.

\begin{figure*}
\begin{center}
\includegraphics[width=0.55\textwidth]{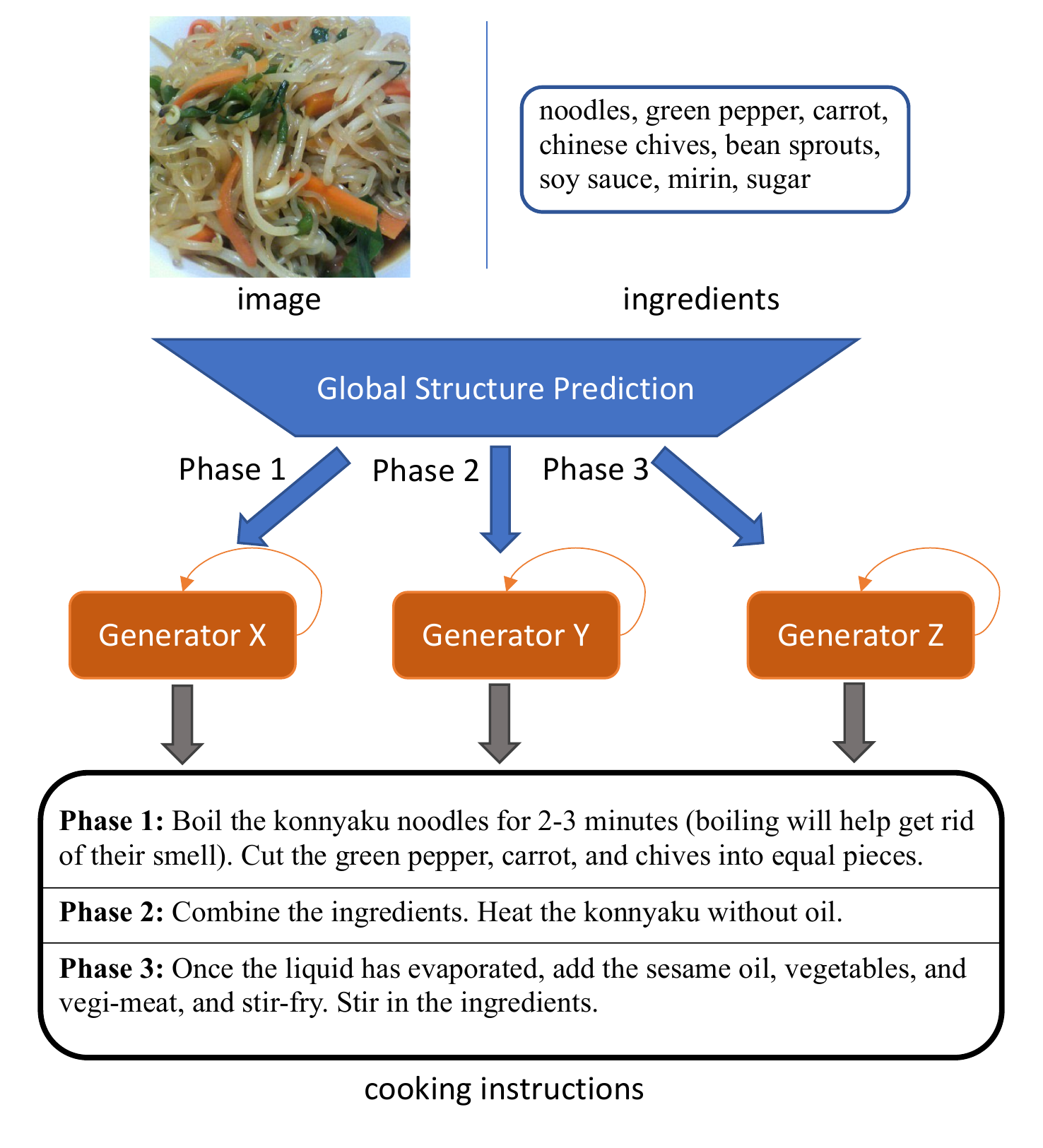}
\end{center}
   \caption{Illustration of the Decomposed Generation Networks (DGN) for recipe generation. Instead of producing instructions directly from the image and ingredient embedding \cite{salvador2019inverse}, we first predict the instruction structure and choose different generators to match the cooking phases. And then we combine the outputs of selected sub-generators to get the final generated recipes.}
\label{fig:DGN_demo}
\end{figure*}

Cooking instructions are one kind of procedural text, which are constructed step by step with some format. For example, as is shown in Figure \ref{fig:DGN_demo}, the cooking instructions are composed of several sentences, and each sentence starts with a verb in most cases. Apart from dividing the cooking instructions by sentences, we may also split them into more general \emph{phases}, which represent the global structures of the cooking recipes. Imagine when people start cooking food, we may decompose the cooking procedure into some basic \emph{phases} first, e.g. \emph{pre-process the ingredients}, \emph{cook the main dish}, etc. Then we will focus on some details, like determining which ingredients to use. While this coarse-to-fine reasoning is trivial for humans, most algorithms do not have the capacity to reason about the phase information contained in the static food image \cite{bosselut2017simulating}. Therefore, it is important to guide the model to be aware of the global structure of the recipe during generation, otherwise the generation outputs can hardly cover all the cooking details \cite{salvador2019inverse}.

Recently, several food datasets have been proposed for recipe generation, such as YouCook2 \cite{zhou2018towards}, Storyboarding \cite{chandu2019storyboarding} and Recipe1M \cite{salvador2017learning}. The first two datasets both include the image sequence, along with their corresponding textual descriptions. The image sequence is a concise series of unfolded cooking videos. Hence the model can obtain the explicit instruction structures with the image sequence. By contrast, Recipe1M remains more challenging, since it only contains the static cooked food images. It is hard to obtain large-scale instructional video data in read world, and sometimes we want to know the exact recipe of a cooked food image. Therefore, we believe that generating cooking instructions from one single food image is of more value, compared to producing instructions from image sequence.

Given the previous stated reasons, we choose the large-scale Recipe1M dataset \cite{salvador2017learning} to implement our methods. Here, our goal is to capture the global structure of recipe and to generate the instruction from one single image with a list of ingredients. The basic idea is that we first (i) assemble some of the consecutive steps to form a \emph{phase}, (ii) assign suitable sub-generators to produce certain instruction phases, and (iii) concatenate the phases together to form the final recipes. We propose a novel framework of \emph{Decomposed Generation Networks} (DGN) with global structure prediction, to achieve the coarse-to-fine reasoning. Figure \ref{fig:framework} shows the pipeline of the framework. To be specific, DGN is composed of two components, i.e. the global structure prediction component and the sub-generator output component. To obtain the global structure of the cooking instruction, we input image and ingredient representations into global structure prediction component, and get the sub-generator selections as well as their orders. Then in the sub-generator output component, we adopt attention mechanism to get the phase-aware features. The phase-aware features are designed for different sub-generators and help the sub-generators produce better instruction phases.

We have conducted extensive experiments on the large-scale Recipe1M dataset, and evaluated the recipe generation results by different evaluation metrics. We find our proposed model DGN outperforms the state-of-the-art methods.

\section{Related Work}

\subsection{Food Computing}
Our work is closely related to food computing \cite{min2018survey}, which utilizes computational methods to analyze the food data including the food images and recipes. With the development of social media and mobile devices, more and more food data become available on the Internet, the UEC Food100 dataset \cite{matsuda2012recognition} and ETHZ Food-101 dataset \cite{bossard2014food} are proposed for the food recognition task. The previous two food datasets are restricted to the variety of data types, only have different categories of food images. YouCook2 dataset is proposed by Zhou et al. in \cite{zhou2018towards}, which contains cooking video data. They focused on generating cooking instruction steps from video segments in YouCook2 dataset. The latter work \cite{chandu2019storyboarding} proposed a new food dataset, Storyboarding, where the food data item has multiple images aligned with instruction steps. In their work, they proposed to utilize a scaffolding structure for the model representations.
Besides, Bosselut et al. \cite{bosselut2017simulating} generated the recipes based on the text, where they reasoned about causal effects that are not mentioned in the surface strings, they achieved this with memory architectures by dynamic entity tracking and obtained a better understanding on procedural text.

In order to better model the relationship between recipes and food images, Recipe1M \cite{salvador2017learning} has been proposed to provide richer food image, cooking instruction, ingredient, and semantic food-class information. Recipe1M contains large amounts of image-recipe pairs, which can be applied on cross-modal food retrieval task \cite{salvador2017learning,carvalho2018cross,wang2019learning} and recipe generation task \cite{salvador2019inverse}. Salvador et al. \cite{salvador2019inverse} focused more on the ingredient prediction task. For instruction generation, they generated the whole cooking instructions from given food images and ingredients through a single decoder directly, which may result in that some cooking details can be missing in some cases. 

It is worth noting that, to the best of our knowledge, \cite{salvador2019inverse} is the only work for recipe generation task on Recipe1M dataset. Our DGN approach improves the recipe generation performance by introducing the decomposing idea to the generation process. Therefore, our proposed methods can be applied to many general models. We will demonstrate the details in Section \ref{exp}.

\subsection{Text Generation}
Text generation is a widely researched task, which can take various input types as source information. Machine translation \cite{sutskever2014sequence,vaswani2017attention} is one of the representative works of text-based generation, in which the decoder takes one language text as the input and outputs another language sentences. Image-based text generation involves both vision and language, such as image captioning \cite{xu2019multi,yang2018multitask,xiao2019deep}, visual question answering \cite{yu2020reasoning,huasong2020self}. To be specific, image captioning is to generate suitable descriptions for the given images, and the goal of visual question answering is to answer questions accompanied with the image and text. In this paper, we try to address the challenging recipe generation problem, which produces a long procedural text conditioned on the image and text (ingredients).

Text generation related tasks are accelerated by some new state-of-the-art models like the Transformer \cite{vaswani2017attention} and BERT \cite{devlin2018bert}, which are attention-based. Many recent works achieve superior performance with attention-based models \cite{yang2019learning,rennie2017self,zha2019context}. In our work, we compare the results of using the pre-trained BERT \cite{devlin2018bert} and normal embedding layer \cite{salvador2019inverse} as the ingredient encoder.

\subsection{Neural Module Networks}
The idea of using neural module network to decompose neural models have been proposed for some language-vision intersection tasks, such as visual question answering \cite{andreas2016neural}, image captioning \cite{yang2019learning}, visual reasoning \cite{hudson2018compositional}. Neural module network has good capabilities to capture the structured knowledge representations of input images or sentences. In general, since the image layouts or questions are obviously structured, many prior related research \cite{andreas2016neural,yang2019learning,hudson2018compositional}, focused on constructing better encoders with neural modules. To produce a coherent story for an image in MS COCO \cite{lin2014microsoft}, Krause et al. \cite{krause2017hierarchical} decomposed both images and paragraphs into their constituent parts, detecting semantic regions in images and using a hierarchical recurrent neural network to generate topic vectors with their corresponding sentences, but they generate different paragraph parts with the same decoder.

In food data \cite{salvador2017learning}, the cooking instructions tend to be very structured as well. To generate recipes with better structures, we employ different sub-generators to produce different phases of cooking instructions. 


\begin{figure*}[htb]
\begin{center}
\includegraphics[width=\textwidth]{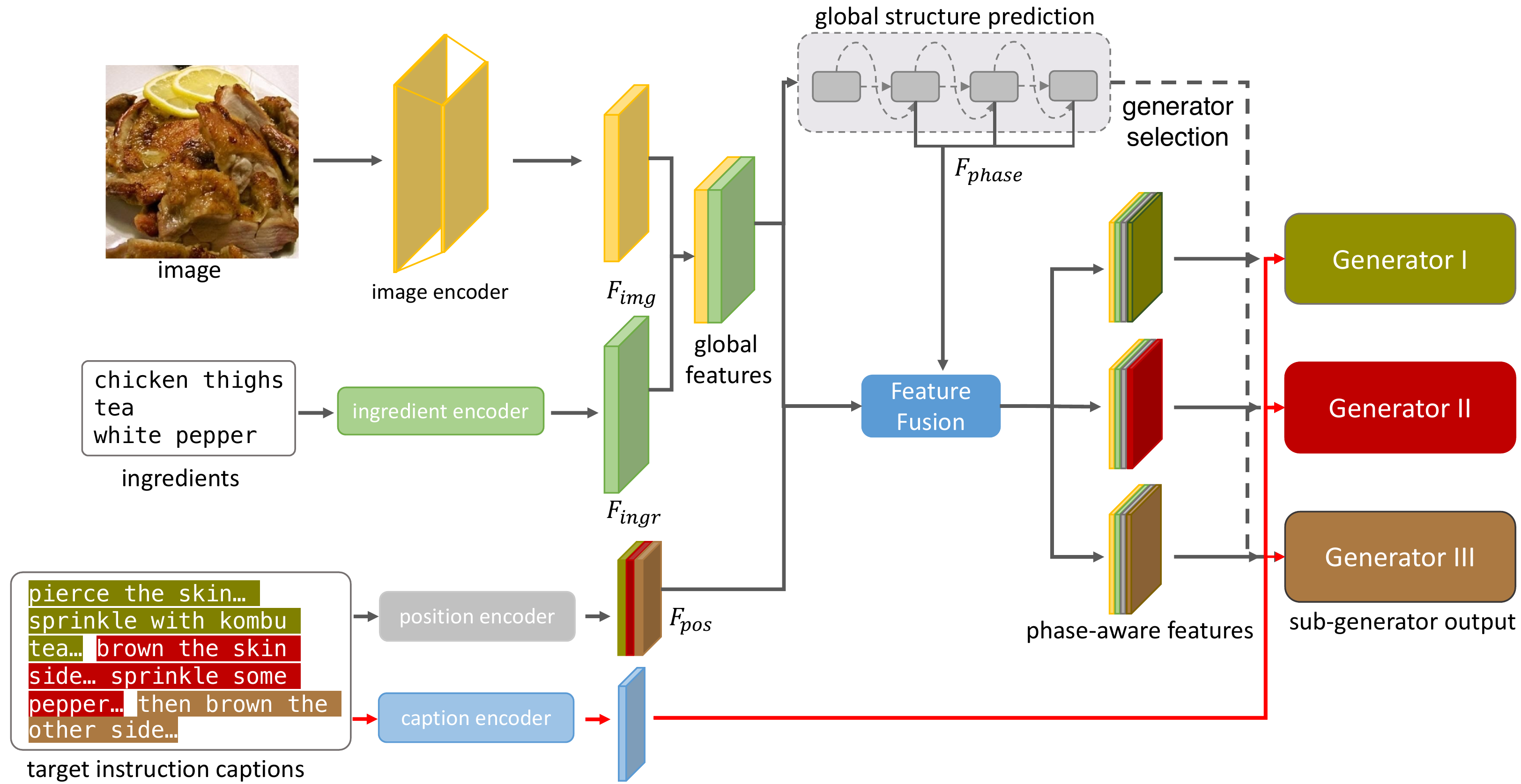}
\end{center}
   \caption{\textbf{Decomposed Generation Networks with global structure prediction (DGN):} We take food images and the corresponding ingredients as model inputs, and obtain the image and ingredient embedding $F_{img}$, $F_{ingr}$ through a pre-trained image model CNN and the language model BERT respectively. After that, the model will be split into two branches, i.e. the global structure prediction component and the sub-generator output component. Both of them are constructed by the transformer. The global structure prediction component produces the sub-generator selections and their orders for the following branch. The sub-generator output component fuses $F_{img}$, $F_{ingr}$, the position representations $F_{pos}$ and the phase vector $F_{phase}$ to obtain the input of each sub-generator, and produces different phases of the recipe.}
\label{fig:framework}
\end{figure*}

\section{Method}
\subsection{Overview}
In Figure \ref{fig:framework}, we show the training flow of DGN. It is observed that the cooking instructions have obvious structures and clear formats, most cooking instruction sentences in Recipe1M dataset \cite{salvador2017learning} start with a verb, e.g. \emph{heat}, \emph{combine}, \emph{pierce}, etc. However, how to automatically divide the recipes into phases remains an NLP problem. Therefore, we use a pre-defined rule to segment the recipes. Specifically, we split per instruction into $2$-$3$ phases and try to ensure each phase shares equal sentence numbers, where one or more cooking steps (sentences) will map to one phase. This recipe segmentation rule is based on intuitions, i.e. having more recipe phases may result in looser cooking step clustering and consequently fail to form the hierarchy between cooking phase and step. 
As an example stated in Figure \ref{fig:framework}, the recipe for the \emph{roasted chicken} totally has five steps, which are transitioned to three phases.

After we obtained the phase segmentation in recipes, we need to determine which sub-generators will be selected to generate the certain phases. We use the approach of k-means clustering to assign pseudo labels to each recipe phase. Specifically, we first extract all the verbs in recipes with spaCy \cite{spacy2}, a Natural Language Processing (NLP) tool. Then, we can obtain the mean verb representations, which can be regarded as the representation of each phase. After that, we use k-means clustering to get pseudo labels for phases, which indicate the selections of sub-generators. The number of the sub-generator category $N$ is a hyper-parameter, we do experiments with different $N$ and show the results in Table \ref{tab:N_impact}. The pseudo labels $Generators = \{g_i, ..., g_k\}$ represent different sub-generator selections.

Figure \ref{fig:framework} provides an overview of our proposed model, which is composed of the global structure prediction component and sub-generator output component. Our model takes food images and their corresponding ingredients as input. It uses several sub-generators for different recipe phases, allowing sub-generators to focus on different clustered recipe phases.

ResNet-50 \cite{he2016deep} pretrained on ImageNet \cite{deng2009imagenet} and BERT \cite{devlin2018bert} model implemented by \cite{Wolf2019HuggingFacesTS} are used to encode food images and ingredients respectively. We can get image and ingredient global representations $F_{img}$ and $F_{ingr}$. These global representations will be fed into the global structure prediction component, to decide which sub-generators will be selected as well as their orders. To enable the interactions among sub-generators, the global structure prediction component also produces a $P$-dimensional phase vector $F_{phase}$ for each of the sub-generators. Then we split the target instructions into phases and assign different position one-hot vectors $v_p \in \mathbb{R}^3$ for each phase, which will be transformed into a $P$-dimensional position representations $F_{pos}$ through a linear layer. With previous encoded features $F_{img}$, $F_{ingr}$, $F_{phase}$ and $F_{pos}$, we can fuse them together and obtain the phase-aware features $\mathbf{r_i} \in \mathbb{R}^P$ for sub-generator $g_i$.

\subsection{Global Structure Prediction Component}\label{structure}
Since the cooking instructions are divided into phases, the global structure prediction component not only needs to decide which generators to be selected in each phase, but also is required to predict the order of the chosen sub-generators. In order to achieve the goal, we stack the transformer blocks \cite{vaswani2017attention} to construct our global structure prediction component. The last transformer block is followed by a linear layer and a softmax activation, to find the predictions for each step. We set hidden size $H=512$, the number of heads $n_{head}=8$ and the number of stacked layers $n_{layer}=4$, generate the sub-generator label sequence $\{y_i, ..., y_k\}$.

To be specific, the transformer block contains two sub-layers with layer normalization, where the first one employs the multi-head self-attention mechanism and the second one attends to the model conditional inputs to enhance the self-attention output. The attention outputs can be computed as \cite{vaswani2017attention}, 

\begin{equation}
\begin{aligned}
\mathrm{Attention}(Q, K, V) = \mathrm{softmax}(\frac{QK^T}{\sqrt{d_k}})V,
\end{aligned}
\end{equation}

where the input comes from queries $Q$ and keys $K$ of dimension $d_k$, and values $V$ of dimension $d_v$. We also adopt the multi-head attention mechanism \cite{vaswani2017attention}, which linearly maps $Q, K, V$ with different, learned projections. These different projected results will be concatenated together and get better output values.

\begin{equation}
\begin{aligned}
\mathrm{MultiHead}(Q, K, V) &= \mathrm{Concat}(\mathrm{head_1}, ..., \mathrm{head_h})W^O, \\
\mathrm{head_i} &= \mathrm{Attention}(QW^Q_i, KW^K_i, VW^V_i)\\
\end{aligned}
\end{equation}

Where the projections are matrices $W^Q_i \in \mathbb{R}^{d_k}$, $W^K_i \in \mathbb{R}^{d_k}$, $W^V_i \in \mathbb{R}^{d_v}$ and $W^O \in \mathbb{R}^{d_vn_{head}}$.

We take the global context vectors $\{F_{img}, F_{ingr}\}$ and target recipe phase labels $g = \{[START], g_i, ..., g_k\}$ as inputs when training the model. We first map the discrete labels to a sequence of continuous representations $Z$. The model generates an output sequence $\{y_i, ..., y_k\}$ one element at a time. The target sequence embedding $Z$ will be first fed into the model and processed with multi-head self-attention layers, as follows:

\begin{equation}
\begin{aligned}
\mathrm{H_{self}^{attn}} = \mathrm{MultiHead}(Z, Z, Z),
\end{aligned}
\end{equation}

We further concatenate the context vectors $\{F_{img}, F_{ingr}\}$ together, get the conditional vector $F_{kv}$, which will be attended to refine previous self-attention outputs $\mathrm{H_{self}^{attn}}$, which is defined as:

\begin{equation}
\begin{aligned}
\mathrm{H_{cond}^{attn}} = \mathrm{MultiHead}(H_{self}^{attn}, F_{kv}, F_{kv}),
\end{aligned}
\end{equation}

$\mathrm{H_{cond}^{attn}}$ is the final attention outputs of each phase, which can be used as the phase vector $F_{phase}$ for sub-generator output component. We transform $\mathrm{H_{cond}^{attn}}$ into $\mathrm{H_{cond}^{attn}}'$ for output token generation with a linear layer. The dimension of $\mathrm{H_{cond}^{attn}}'$ is identical with the number of sub-generator category $N$, the probabilities of generated tokens are $p^{gen} = \mathrm{softmax}(\mathrm{H_{cond}^{attn}}')$. Therefore, the final output tokens of global structure prediction component $y_i = \mathrm{argmax}(p^{gen})$. We train the global structure prediction component with cross-entropy loss $\mathcal{L}_{pre}$:

\begin{equation}
\begin{aligned}
\mathcal{L}_{pre} = \sum_{i=1}^S\ell_{cross-entropy}(p_{i}^{gen}, g_{i}),
\end{aligned}
\end{equation}

where $S$ is the number of instruction phases.

\subsection{Sub-Generator Output Component}

The sub-generator output component uses different sub-generators predicted by global structure prediction component, to produce a certain phase of the recipe, and concatenate them together to form the final cooking instruction. We stack $16$ transformer blocks to construct the generator, in which $12$ of them are shared blocks, and the rest $4$ are independent blocks of each of the generators. The reasons for using shared blocks lie in that the model may overfit to the limited training data and cannot generalize well, if we adopt whole independent blocks for each sub-generator.

We utilize each predicted sub-generator to produce one recipe phase, which requires that each of the generator inputs should be discriminative and informative enough. Therefore, we incorporate rich sources of feature representations, i.e. the food image features $F_{img}$, the ingredient features $F_{ingr}$, the position representations $F_{pos}$ and the phase vector $F_{phase}$ ($\mathrm{H_{cond}^{attn}}$) produced by global structure prediction component. $F_{img}$ provides the model with generation contents from the food images, which belong to a different modality, and $F_{ingr}$ indicates the ingredients containing in the recipe, which can be reused in the generated cooking instructions. To allow the model to be aware of the generation phase, we fuse the recipe phase position representations $F_{pos}$. $F_{phase}$ is incorporated for enhancing the interactions among different sub-generators and helps the model adapt to different generation phases. 

The above four representations will be fused together to get the phase-aware features $\mathbf{r} = \langle F_{img}, F_{ingr}, F_{pos}, F_{phase} \rangle$, which are the inputs of sub-generators. We adopt two different ways to achieve that. The first one is that we simply concatenate these representations, and get $\mathbf{r_{cat}}$. In the second way, we use attention mechanism to make $F_{img}$, $F_{ingr}$ attend to the concatenated embedding $\mathbf{cat}(F_{pos}, F_{phase})$ respectively. Specifically, we utilize a projection matrix on $\mathbf{cat}(F_{pos}, F_{phase})$ and get the attention maps for $F_{img}$ and $F_{ingr}$, the image and ingredient attention outputs can be formulated as:

\begin{equation}
\begin{aligned}
F_{img}^{attn} = \mathrm{softmax}(\mathrm{W_1}(\mathbf{cat}(F_{pos}, F_{phase})))F_{img}, \\
F_{ingr}^{attn} = \mathrm{softmax}(\mathrm{W_2}(\mathbf{cat}(F_{pos}, F_{phase})))F_{ingr}, 
\end{aligned}
\end{equation}

The final attended phase-aware features $\mathbf{r_{attn}}$ is the concatenation of $F_{img}^{attn}$ and $F_{ingr}^{attn}$. We involve an additional position classifier $\mathcal{L}_{pos}$ on $\mathbf{r}$ to ensure that it contains certain phase position information.

We also need to input the target instruction captions $t=\{[START], t_1, t_2, ..., t_m\}$ for training the Transformer \cite{vaswani2017attention} generators, and map them to a continuous representation $C$. As described in Section \ref{structure}, we utilize attention mechanism with transformer blocks:

\begin{equation}
\begin{aligned}
\mathbf{F}_{self}^{attn} = \mathrm{MultiHead}(C, C, C), 
\end{aligned}
\end{equation}
\begin{equation}
\begin{aligned}
\mathbf{F}_{cond}^{attn} = \mathrm{MultiHead}(\mathbf{F}_{self}^{attn}, \mathbf{r}, \mathbf{r}),
\end{aligned}
\end{equation}

We use $\mathbf{F}_{cond}^{attn}$ to generate the tokens through a linear layer and softmax activation, and we can obtain the output probabilities $p^{token}$ among candidate tokens. For each sub-generator, we compute the training loss as follows:

\begin{equation}
\begin{aligned}
\mathcal{L}_{gen} = \sum_{i=1}^M\ell_{cross-entropy}(p_{i}^{token}, t_{i}),
\end{aligned}
\end{equation}

\subsection{Training and Inference}

The food images, ingredients and the target instruction captions are taken as the training input of the model. We totally have three loss functions, i.e. the global structure prediction loss $\mathcal{L}_{pre}$, sub-generator output loss $\mathcal{L}_{gen}$ and position classification loss $\mathcal{L}_{pos}$, our training loss can be formulated as:

\begin{equation}
\begin{aligned}
\label{eq:main}
\mathcal{L} = \lambda_{1} \mathcal{L}_{pre} + \lambda_{2} \mathcal{L}_{gen} + \lambda_{3} \mathcal{L}_{pos},
\end{aligned}
\end{equation}

The Transformer model \cite{vaswani2017attention} is auto-regressive, which utilizes the previously generated tokens as additional input while generating the next \cite{vaswani2017attention}. Therefore, during inference time, we first feed the model with the $[START]$ token instead of the whole target instruction captions, and then the model will output the following tokens incrementally. We run the global structure prediction component first. According to the predicted sub-generator sequence, we utilize the chosen generator for each recipe phase.

\section{Experiments}\label{exp}

\subsection{Dataset and Evaluation Metrics}
\noindent \textbf{Dataset.} We use the Recipe1M \cite{salvador2017learning,salvador2019inverse} provided official split: $252,547$,
$54,255$ and $54,506$ recipes for training, validation and test respectively. Theses recipes are scraped from cooking websites, and each of them contains the food image, a list of ingredients and the cooking instructions. Since Recipe1M data is uploaded by users, there have large variance and noises across the food images and recipes.

\noindent \textbf{Evaluation Metrics.} We totally adopt three different metrics for evaluation, i.e. perplexity, BLEU \cite{papineni2002bleu}, ROUGE \cite{lin2004rouge}. The prior work \cite{salvador2019inverse} only used perplexity for evaluation, which measures how well the probability distribution of learned words matches that of the input instructions. BLEU scores are based on an average of unigram, bigram, trigram and 4-gram precision, however, it fails to consider sentence structures \cite{callison2006re}. In other words, BLEU cannot evaluate the performance of our global structure prediction component. ROUGE is a modification of BLEU that focuses on recall rather than precision, i.e. it looks at how many n-grams of the reference text show up in the outputs, rather than the reverse. Therefore, ROUGE can reflect the influence of the proposed global structure prediction component, which is discussed in Section \ref{ablation}.

\begin{figure*}[t]
\begin{center}
\includegraphics[width=0.6\textwidth]{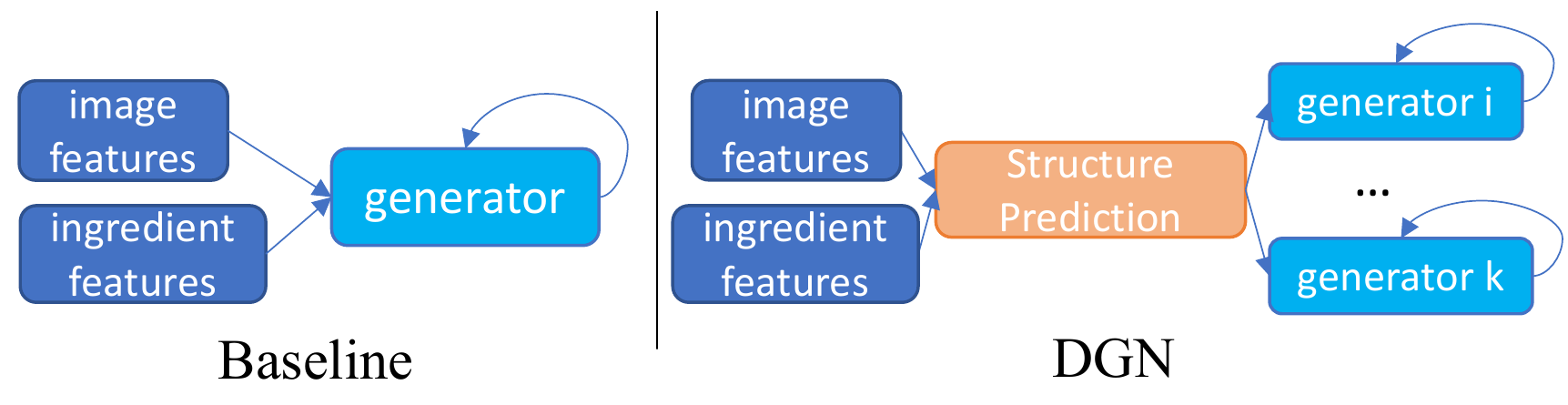}
\end{center}
   \caption{The comparison of the baseline model and our proposed DGN. DGN can be applied to different backbone networks.}
\label{fig:compa}
\end{figure*}

\subsection{Implementation Details}
We utilize ResNet-50 \cite{he2016deep} which is pretrained on ImageNet \cite{deng2009imagenet} as the image encoder, which takes image size of $224 \times 224$ as input. The ingredient encoder is BERT \cite{devlin2018bert}, short for Bidirectional Encoder Representations from Transformers, which is a pretrained language model implemented by \cite{Wolf2019HuggingFacesTS} and is one of the state-of-the-art NLP models. As the prior work setting \cite{salvador2019inverse}, we adopt the last convolutional layer of ResNet-50, whose output dimension is $512$, as the feature representations. \cite{salvador2019inverse} used $20$ ingredients per recipe for embedding, but since BERT tokenizer \cite{devlin2018bert} may split one word into several tokens, so we set the maximum number of tokens as $30$. The output embedding of BERT model will be mapped to the dimension of $512$ as well. For the cooking instruction generators, different sub-generators will share $12$ transformer blocks, and each of them has additional independent $4$ transformer blocks with $8$ multi-head attention heads. To align with \cite{salvador2019inverse} and achieve a fair comparison, we generation instruction of maximum $150$ words. In all the experiments, we use greedy search for recipe generation. 

Regarding the phase number setting of each cooking instruction, we experiment with different numbers and observe that splitting per instruction into up-to three phases has the best trade-off performance. Since the cooking step numbers range from $2$ to $19$, suppose that if we split too many phases for each recipe, one phase may only contain one step, which will fail to obtain the global structure information. Therefore, we assume per instruction has at most three phases.

In all the experiments, we fix the weights of the image encoder for faster training, and instead of using the predicted ingredients as conditional generator inputs \cite{salvador2019inverse}, we take the ground truth ingredients and images as input for a fair comparison. We set $\lambda_1$, $\lambda_2$ and $\lambda_3$ in Eq. \ref{eq:main} to be $1$, $1$ and $0.1$ respectively, which is based on empirical observations on validation set. The model is optimized with Adam \cite{kingma2014adam}, and the initial learning rate is set as $0.001$, with $0.99$ decay per epoch. The model is trained for about $25$ epochs to be converged. We implement the proposed methods with PyTorch \cite{paszke2017automatic}.

\subsection{Baselines}
To the best of our knowledge, \cite{salvador2019inverse} is the only work for recipe generation task at Recipe1M dataset, where they generated the whole cooking instructions from the cooked food images through $16$ transformer blocks. By contrast, our proposed DGN extends an additional branch for the text generation process, which predicts the structures of the recipes first and then utilizes the chosen sub-generators for each phase generation. In other words, DGN can be applied to different backbone networks. We compare the difference between baseline models and the proposed DGN in Figure \ref{fig:compa}.

To fully demonstrate the efficacy of DGN, we experiment with two different ingredient encoders to act as baseline results. The first one comes from the prior work \cite{salvador2019inverse}, where they adopted one word embedding layer to encode the ingredients. We need to train it from scratch. For comparison, BERT \cite{devlin2018bert} is utilized as the second ingredient encoder. We finetune the BERT model during training. Note that the above two baseline models both use ResNet-50 as the image encoder, they only differ in the ingredient encoders.

\begin{table*}[t]
  \centering
  \caption{Main Results. Evaluation of DGN performance against different settings. We first show the results of two ingredient encoders, where the first one adopts the word embedding layer to encode the ingredients, while the second one, BERT, uses a pretrained language model. 
  DGN is added to the baseline models as an additional branch, where we show the results of different construction ways of phase-aware features $\mathbf{r}$. \textbf{DGN (cat)} uses the concatenation of the provided representations for the sub-generator inputs, and \textbf{DGN (attn)} adopts the attention mechanism to enhance the representations.
  We evaluate the model with perplexity (lower is better), BLEU (higher is better) and ROUGE-L (higher is better). We find the proposed DGN improves the performance across all the metrics.}
  {
    \begin{tabular}{llccc}
    \toprule
    \cline{1-5}
     \textbf{Methods} & \textbf{Ingredient Encoder} & \textbf{Perplexity} & \textbf{BLEU} &\textbf{ROUGE-L} \\
    \midrule
    \textbf{Baseline} \cite{salvador2019inverse} & \textbf{Embedding Layer} & 8.06  & 7.23  & 31.8  \\
    \textbf{DGN (cat)} & \textbf{Embedding Layer} &   7.40  & 9.93  & 34.5 \\
    \textbf{DGN (attn)} & \textbf{Embedding Layer} &  7.34  & 10.51  & 34.9 \\
    \midrule
    \textbf{Baseline} \cite{devlin2018bert} & \textbf{BERT} & 7.52 & 9.29  & 34.8 \\
    \textbf{DGN (cat)} & \textbf{BERT} &  6.78 & 10.76  & 36.0 \\
    \textbf{DGN (attn)} & \textbf{BERT} & \textbf{6.59} & \textbf{11.83}  & \textbf{36.6} \\
    \cline{1-5}
    \bottomrule
    \end{tabular}%
  }
  \label{tab:main}%
\end{table*}%

\subsection{Main Results}
We show our main results of generating cooking instructions in Table \ref{tab:main}, which are evaluated across three language metrics: perplexity, BLEU \cite{papineni2002bleu} and ROUGE-L \cite{lin2004rouge}. Generally, models with and without DGN have an obvious performance gap. Simply using one word embedding layer for ingredient encoder performs poorly, achieving the lowest scores across all the metrics. When we replace the embedding layer with state-of-the-art pretrained language model, BERT, the performance reasonably gets better, which highlights the significance of the pretrained model.

We then incrementally add the DGN branch to two different backbone networks. To be specific, we experiment with two ways to construct the phase-aware features $\mathbf{r}$, i.e. \textbf{DGN (cat)}, where $\mathbf{r}$ is formed by the concatenation of the four representations, and \textbf{DGN (attn)}, in which we construct image and ingredient features with attention mechanism, then concatenate them together to be $\mathbf{r}$. First, we add \textbf{DGN (cat)} to baseline models, surprisingly this approach can achieve more than $2$ BLEU scores better than the baseline model with embedding layer and $1$ BLEU score over state-of-the-art language model BERT, which indicates our DGN idea is very promising and can extend to some general models. We further adopt \textbf{DGN (attn)} for recipe generation evaluation, the performance continually gets better, illustrating the usefulness of enhancing the inputs of generators. In general, our full model, \textbf{BERT + DGN (attn)}, obtains the best results among all methods on every metric consistently, and achieve the state-of-the-art performance.

\begin{table*}
  \centering
  \caption{The ablative influence of image and ingredient as input. The model is evaluated by perplexity (lower is better), BLEU (higher is better) and ROUGE-L (higher is better).}
  {
    \begin{tabular}{lcccc}
    \toprule
    \midrule
    \textbf{Input} & \textbf{Perplexity} & \textbf{BLEU} &\textbf{ROUGE-L} \\
    \midrule
    \textbf{Only Image} & 8.16  & 3.72  & 31.0 \\
    \textbf{Only Ingredient} & 7.62 & 5.74  & 32.1 \\
    \textbf{Image and Ingredient} & \textbf{7.52}  &	\textbf{9.29}  & \textbf{34.8}  \\
    \midrule
    \bottomrule
    \end{tabular}%
  }
  \label{tab:img_ingr}%
\end{table*}%

\begin{table*}
  \centering
  \caption{The impact of sub-generator category number $N$. The model is evaluated by perplexity (lower is better), BLEU (higher is better) and ROUGE-L (higher is better).}
  {
    \begin{tabular}{lcccc}
    \toprule
    \midrule
    \textbf{N} & \textbf{Methods} & \textbf{Perplexity} & \textbf{BLEU} &\textbf{ROUGE-L} \\
    \midrule
    \textbf{1} & \textbf{BERT} & 7.52 & 9.29  & 34.8 \\
    \midrule
    \textbf{1} & \textbf{BERT+DGN} & 6.98  & 10.98  & 35.8 \\
    \textbf{3} & \textbf{BERT+DGN} & \textbf{6.59} & \textbf{11.83}  & \textbf{36.6} \\
    \textbf{5} & \textbf{BERT+DGN} & 6.95  & 11.15  & 36.0  \\
    \midrule
    \bottomrule
    \end{tabular}%
  }
  \label{tab:N_impact}%
\end{table*}%

\begin{table*}
  \centering
  \caption{The impacts of DGN on the average length and vocabulary size of generated recipes. The results demonstrate that the proposed DGN increases the average length and diversity of generated cooking instructions.}
  {
    \begin{tabular}{lcc}
    \toprule
    \midrule
    \textbf{Methods} & \textbf{Average Length} & \textbf{Vocab Size}\\
    \midrule
    \textbf{Baseline} \cite{salvador2019inverse} &  69.9 & 3657 \\
    \textbf{Baseline} \cite{devlin2018bert} & 66.9 & 4521  \\
    \textbf{DGN (Baseline \cite{salvador2019inverse})} & 103.1  & 4836 \\
    \textbf{DGN (Baseline \cite{devlin2018bert})} & 105.6 & 6573 \\
    \midrule
    \textbf{Ground Truth} & 116.5 & 33110 \\
    \midrule
    \bottomrule
    \end{tabular}%
  }
  \label{tab:ave_len}%
\end{table*}%

\subsection{Ablation Studies}\label{ablation}
\noindent \textbf{The ablative influence of image and ingredient as input.} To suggest the necessity of using both image and ingredient as input, we train the model with different inputs separately. We show the ablation studies in Table \ref{tab:img_ingr}, where we use a transformer for generation, instead of DGN. It can be observed that ingredient information helps more on the recipe generation, since ingredients can be directly reflected in the recipes. The model with image and ingredient as input has better performance than that of single modality input.

\noindent \textbf{The impact of sub-generator category number $N$.} 
After we get the representation of each instruction phase, we adopt k-means clustering to obtain the phase labels, which indicate the sub-generator selections. Then these labels are used for the global structure prediction component training. 
We show the experiment results in Table \ref{tab:N_impact}, where the first row shows the experiment results of BERT baseline model, the last four rows are all implemented by BERT + DGN (attn). When $N=1$, we compare the results of the first and second row, the first row uses the concatenated representations of image and ingredient features, while the second row takes the enhanced phase-aware features $\mathbf{r}$ as input, indicating the efficacy of the phase-aware features.
Besides, the model with $N=1$ has inferior performance compared with model with $N=3$, illustrating the single generator struggles to fit data from different phases. When $N=5$, the model gets similar evaluation results to $N=1$. That model with $N=5$ has poorer performance than model with $N=3$ may because the model does not have enough data for training, due to the more splits of the training data.
Therefore, we set the hyper-parameter $N$ to be $3$.

\begin{figure*}[t]
\begin{center}
\includegraphics[width=1\textwidth]{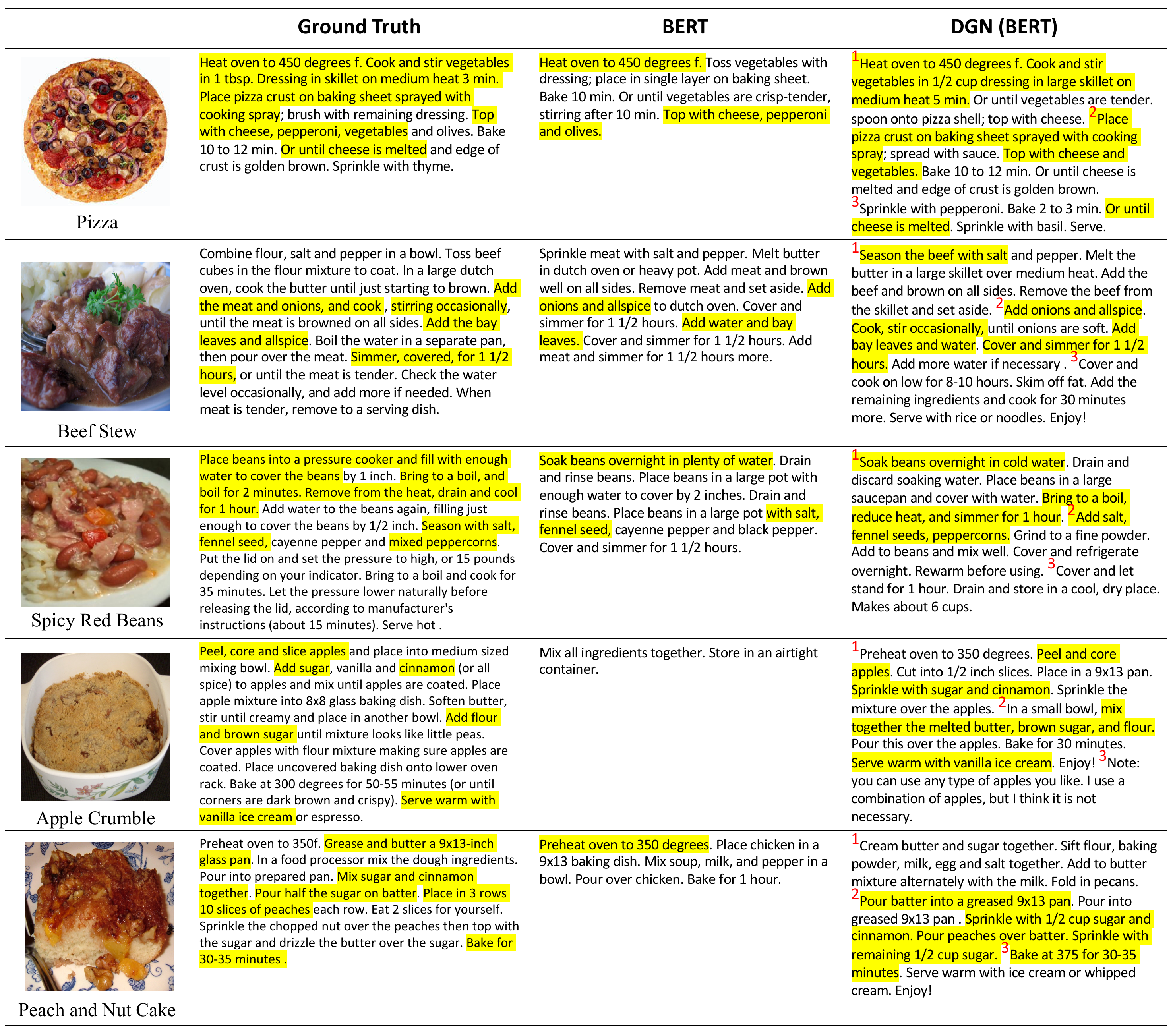}
\end{center}
   \caption{Analysis of generated recipes by different models. We show the generated results conditioned on three different food images, namely \emph{pizza}, \emph{beef stew} \emph{spicy red beans}, \emph{apple crumble} and \emph{peach and nut cake}. The left column shows the conditional food images, and the right three columns show the true cooking instructions, baseline BERT generations and DGN generated recipes. Words with yellow background represent the matching parts between raw recipes and the generated recipes. In the DGN generations, we state the recipe phases with numbers in red.}
\label{fig:re_gen}
\end{figure*}

\noindent \textbf{The impacts of DGN on the average length and vocabulary size of generated recipes.} In order to further demonstrate the effectiveness of the proposed DGN from other aspects, we perform some language analysis based on the generated outputs in the Table \ref{tab:ave_len}. Our DGN approach generates text of the closet average length as ground truth recipes, which are crawled from websites and written by humans. While the models without DGN generate relatively short cooking instructions, which provides the evidence for our assumptions before: using one single generator will result in some cooking details are missing. We also show some qualitative results in Figure \ref{fig:re_gen}. To evaluate the diversity of the recipes, we compute the vocabulary sizes of the generations and the ground truth, which indicates the number of unique words that appear in the text. According to the results, DGN (BERT) is actually the most diverse method apart from the ground truth. But there still remain huge gaps between the diversity of generated text and human-written text.

\noindent \textbf{The effect of global structure prediction.} Global structure prediction component is the first and basic part of our proposed DGN model, which outputs the sub-generator selections and their orders for subsequent generations. We test the generated text of the predicted orders and that of random orders. We adopt the ROUGE-L metric for evaluation, since BLEU focuses on the recall instead of the precision and it cannot reflect the impact of different orders of recipe phases, while ROUGE considers both recall and precision. The random order output ROUGE-L score turns out to be $0.335$, about $3$ percentage drop from the predicted order evaluation results.

\subsection{Qualitative Results}
We present some qualitative results from our proposed model and the ground truth cooking instructions for comparison in Figure \ref{fig:re_gen}. In the left column, we show the conditional food images, which come from \emph{pizza}, \emph{beef stew}, \emph{spicy red beans}, \emph{apple crumble} and \emph{peach and nut cake} respectively. And in the right three columns, we list the true recipes, the generated recipes of BERT and that of our proposed model DGN, which uses the attended features. We indicate the recipe phases with the red number in DGN generations, and words with yellow background suggest the matching parts between raw recipes and the generated recipes.

The obvious properties of DGN generations include its average length and its ability to capture rich cooking details. First of all, we can see that DGN generates longer recipe outputs than BERT, which has a similar length as true recipes. Besides, it is observed that the phase orders predicted by the global structure prediction component make sense in the shown cases: the first instruction phase gives some instructions on pre-processing the ingredients, the middle instruction phase tends to describe the details about the main dish cooking, and the last phase often contains some concluding work of cooking. 

Generally, it can be seen that DGN generates more matching cooking instruction steps with the ground truth recipes than BERT. When we go into the details, the DGN generated instructions include the ingredients used in the true recipes. Specifically, in the top row, the generated text covers the ingredients of \emph{pepperoni}, \emph{cheese}, \emph{vegetables} and etc. Compared with the BERT outputs, DGN generate similar sentences at the beginning. However, DGN provides more details, e.g. in the instruction generation of \emph{beef stew}, both BERT and DGN output the sentence of ``Add onions and allspice.", while DGN further generate some tips: ``Cook, stir occasionally, until onions are soft.".

It is also worth noting that some of the predicted numbers are not precise enough, like in the third generated phase of \emph{Beef Stew}, the generation output turns out to be ``cook ... for 8-10 hours", which is not aligned with common sense.

\section{Conclusion}
In this paper, we have proposed to make the generated cooking instructions more structured and complete, i.e. to decompose the recipe generation process. In particular, we present a novel framework DGN for recipe generation that leverages the compositional structures of cooking instructions. Specifically, we first predicted the global structures of the instructions based on the conditional food images and ingredients, and determined the sub-generator selections and their orders. Then we constructed a novel phase-aware feature for the input of chosen sub-generators and adopted them to produce the instruction phases, which are concatenated together to obtain the whole cooking instructions. Experimentally, we have demonstrated the advantages of our approach over traditional methods, which use one single decoder to generate the long cooking instructions. We conducted extensive experiments with ablation studies, and achieved state-of-the-art recipe generation results across different metrics in Recipe1M dataset.



\section*{Acknowledgment}

This research is supported, in part, by the National Research Foundation (NRF), Singapore under its AI Singapore Programme (AISG Award No: AISG-GC-2019-003) and under its NRF Investigatorship Programme (NRFI Award No. NRF-NRFI05-2019-0002). Any opinions, findings and conclusions or recommendations expressed in this material are those of the authors and do not reflect the views of National Research Foundation, Singapore. This research is also supported, in part, by the Singapore Ministry of Health under its National Innovation Challenge on Active and Confident Ageing (NIC Project No. MOH/NIC/COG04/2017 and MOH/NIC/HAIG03/2017), and the MOE Tier-1 research grants: RG28/18 (S) and RG22/19 (S).

\newpage

\bibliographystyle{IEEEtran}
\bibliography{IEEEexample}




%








\end{document}